# A Topic Modeling Analysis of Stigma Dimensions, Social, and Related Behavioral Circumstances in Clinical Notes Among Patients with HIV

Authors:

Ziyi Chen, MS[1]
Yiyang Liu, PhD[2]
Mattia Prosperi, PhD[2]
Krishna Vaddiparti, PhD[2]
Robert L Cook, MD[2]
Jiang Bian, PhD[4,5]
Yi Guo, PhD[1]
Yonghui Wu, PhD[1,3]

Affiliation of the authors:

[1]Department of Health Outcomes and Biomedical Informatics, College of Medicine, University of Florida, Gainesville, Florida, USA
[2]Department of Epidemiology, College of Public Health and Health Professions, College of Medicine, University of Florida, Gainesville, Florida, USA

[3]Preston A. Wells, Jr. Center for Brain Tumor Therapy, Lillian S. Wells Department of Neurosurgery, University of Florida, Gainesville, Florida, USA

[4]Department of Biostatistics and Health Data Science, School of Medicine, Indiana University, Indianapolis, Indiana, USA

[5]Regenstrief Institute, Indianapolis, Indiana, USA

Corresponding author:

Yonghui Wu, PhD
1889 Museum Rd, Suite 7000
PO Box 100147
Gainesville, FL, USA, 32611
Phone: 352-294-8436
Email: yonghui.wu@ufl.edu

Keywords:

Human Immunodeficiency Virus
Stigma
Topic Modeling
Natural Language Processing
Electronic Health Records
Clinical Notes

**ABSTRACT**

**Objective**

To characterize stigma dimensions, social, and related behavioral circumstances in people living with HIV (PLWHs) seeking care, using natural language processing methods applied to a large collection of electronic health record (EHR) clinical notes from a large integrated health system in the southeast United States.

**Methods**

We identified a cohort of PLWHs from the University of Florida (UF) Health Integrated Data Repository and performed topic modeling analysis using Latent Dirichlet Allocation (LDA) to uncover stigma-related dimensions and related social and behavioral contexts. Domain experts created a seed list of HIV-related stigma keywords, then applied a snowball strategy to iteratively review notes for additional terms until saturation was reached. To identify more target topics, we tested three keyword-based filtering strategies. The detected topics were evaluated using three widely used metrics and manually reviewed by specialists. Word frequency analysis was used to highlight the prevalent terms associated with each topic. In addition, we conducted topic variation analysis among subgroups to examine differences across age and sex-specific demographics.

**Results**

We identified 9,140 PLWHs at UF Health and collected 2.9 million clinical notes. Through the iterative keyword approach, we generated a list of 91 keywords associated with HIV-related stigma.  Topic modeling on sentences containing at least one keyword uncovered a wide range of topic themes associated with HIV-related stigma, social, and related behaviors circumstances, including “Mental Health Concern and Stigma”, “Social Support and Engagement”, “Limited Healthcare Access and Severe Illness”, “Missed Appointments and HIV Care Monitoring”, “Treatment Refusal and Isolation”, “Intimate Partner Violence and Relationship Concerns”, “Fear of Falling and Physical Health Concerns”, “Substance Abuse”, and “Food Insecurity and Resource Scarcity”.  Topic variation analysis across sex and age subgroups revealed no substantial difference between males and females; however, there were differences were observed among different ages. For example, "Fear of Falling and Physical Health Concerns" was notably more prevalent among older adults.

**Conclusion**

Extracting and understanding the HIV-related stigma and associated social and behavioral circumstances from EHR clinical notes enables scalable, time-efficient assessment and overcoming the limitations of traditional questionnaires. Findings from this research provide actionable insights to inform patient care and interventions to improve HIV-care outcomes.

## 1. INTRODUCTION AND BACKGROUND

The human immunodeficiency virus (HIV) is recognized as the causative agent of acquired immunodeficiency syndrome (AIDS), a global health burden. Based on the World Health Organization's estimates, 91.4 million people have been infected with HIV, and approximately 44.1 million have died from AIDS or related complications since the beginning of the epidemic in the 1980s. As of the end of 2024, an estimated 40.8 million people were living with HIV worldwide [1]. In addition to medical complications, comorbidities, and AIDS progression, people living with HIV (PLWHs) have historically experienced stigma across diverse global contexts, including both internalized stigma, such as disclosure concerns and negative self-image, and external stigma such as discrimination and social stigma [2]–[4]. The stigma can trigger anxiety, depression, and other mental health challenges, undermining treatment adherence and care engagement [5]–[7]. Compounding this, social and related behavioral circumstances, such as discrimination in clinical settings, limited medical and social support, which further discourage HIV-infected people from seeking help, negatively impacting their mental health [8], [9]. Together, these intersecting obstacles adversely affect their mental well-being, physical health outcomes, and overall quality of life. Therefore, assessing and addressing stigma, alongside these social and behavioral circumstances, are crucial for care management of PLWHs and improvement of outcomes, both short and long term.

The traditional way to measure stigma and social, behavioral determinants of health is through surveys and questionnaires [2], [10]–[12], with standardized instruments tailored for PLWHs. For instance, the Berger stigma scale [2] utilizes 40 likert-scale items to assess personalized stigma, concerns about disclosure, negative self-image, and public attitudes. Holzemer [12] developed and tested an instrument to measure perceived stigma, establish a baseline for tracking changes in stigma over time, and monitor progress towards stigma reduction. More recently, studies have looked at social media to examine HIV-related discussions and their social, behavioral aspects, including stigma [13], [14]. Unfortunately, surveys are hard to scale up for large populations, as they are time and resource-consuming, and have a significant burden on the responders. While social media offers a much larger sample size, there is considerable selection bias, and it is extremely difficult to link such to health measurements at the individual level.

There has been an increasing interest in utilizing clinical notes from electronic health records (EHR) of healthcare encounters to study stigma, social, and related behavioral circumstances, since this information is not typically captured in a structured data field but may be documented in clinical notes. Clinical notes collected during routine care capture rich, physician-documented details on patients' experiences and are considered more accurate and reliable than social media. Compared with the traditional survey approaches, EHR notes minimize patient burden by avoiding the additional time and effort required to complete questionnaires, removing the need for staff training on survey administration and participant recruitment, and reducing issues such as low response rates and inconsistent question wording. Currently, these notes have been employed to screen for HIV risk behavior and mental health conditions in PLWHs, but have not been leveraged yet for characterizing HIV-related stigma and its dimensions [11], [15].

Topic modeling is a widely used method to discover hidden topics from large corpora in various fields [16]–[18] and can be a fundamental step to categorizing, understanding HIV-related stigma and social, behavioral factors from clinical notes. Previous studies have applied topic modeling to clinical notes, investigating COVID-19 effects in community health [19], natural history of autoantibodies directed against neutrophil cytoplasmic antigens (ANCA)-associated vasculitis [20], rapid response in nursing interventions [21], and patient trajectories of Alzheimer's disease and related dementias (ADRD) [22].

In this work, we conducted a large-scale, time-efficient investigation of stigma leveraging the real-world EHR. We employed the Latent Dirichlet Allocation (LDA) [23], the most widely used unsupervised probabilistic topic modeling technique [24], to discover the linguistic themes and discourses from PLWHs, focusing on stigma dimensions, social, and related behavioral circumstances, and analyzing differences by demographic groups.

## 2. MATERIALS AND METHODS

### 2.1. Data Sources

The study used EHR, including clinical notes, from PLWHs who have received care between 2012 and 2022 at University of Florida (UF) Health, a large academic medical center in Florida, with several hospitals and satellite clinics covering a rural-urban continuum for over 2 million people. PLWHs were identified using a computable phenotype algorithm [25]. In brief, the algorithm identifies PLWHs if they have at least one HIV diagnosis code and either laboratory records suggesting HIV-positive status, prescription or medication dispense records for HIV treatment, or multiple HIV-related clinical encounters. The algorithm has been validated via chart review, achieving a 98.9% sensitivity and 97.6% specificity.

The UF Institutional Review Board (IRB) approved the study protocol under protocol IRB202300703, which governs our access to the limited clinical notes dataset. Data were obtained from our clinical data warehouse, the UF Integrated Data Repository (IDR), which contains fully digitized patient records from diverse healthcare settings, including but not limited to inpatient, outpatient, and emergency department visits, and extracted from multiple information systems such as Epic. All the protected health information was removed during quality checks. The clinical notes in our dataset include all narrative free-text documents entered by providers into the patient EHR at UF Health, including progress notes, plan of care, admission, and discharge summaries. Scanned documents requiring optical character recognition (OCR) were excluded. The full list of note types was provided in **Supplementary Table 1**.

### 2.2. Data Preprocessing

A total of 2.9 million clinical notes were identified from this HIV cohort. As not all notes contain stigma documentations, we experimented with three strategies to refine clinical notes used for topic modeling, as shown in **Figure 1A**. The first strategy used all clinical notes from PLWHs to conduct topical modeling analysis. The second strategy used clinical notes with at least one stigma-related keyword for topic modeling. The third strategy only used the sentences that contain at least one identified keyword.

To compose the keywords, we asked HIV experts to compile a list of keywords related to stigma based on the questions from established HIV stigma measures, such as the Berger stigma scale. Based on this keyword list, we used a snowball strategy to iteratively review a small batch of randomly sampled notes from the full corpus using a uniform random sampling procedure to identify new keywords until no more new keywords could be identified. To further extend the keyword list, we also identified the synonyms of these keywords by calculating cosine similarity using the word embeddings derived using GatorTron [26], a clinical large language model developed at UF.

We performed common text cleaning and processing of the notes using the NLTK package [27], such as removing unreadable characters, numbers, punctuation, and stop words, from the original text. We used lemmatization to transform words back to their base form to minimize the vocabulary size. All text was converted to lowercase. Duplicates and notes with fewer than 50 characters were removed.

### 2.3. Topic Modeling

LDA is a widely used and well-validated probabilistic topic modeling technique for discovering and identifying potential topics within a corpus. LDA assumes that documents are generated from a mixture of multiple topics, and the probability of words occurring in a specific topic varies among different topics. A document-term matrix is used to calculate topic distributions based on word frequency and co-occurrence.

In this study, we used the Gensim LDA implementation. [28] The LDA model was trained with hyperparameters 'alpha' set to 'auto' and 'beta' set to 'symmetric'. The number of 'passes' over the corpus is equal to 10. The inference parameter 'iterations' was fixed at 10 to control convergence during model training. To ensure the stability and robustness of the generated topics, we conducted 10 iterations through the corpus during the topic modeling analyses with different random seeds. The number of topics (K) is one of the most important hyperparameters for LDA analysis. To determine the optimal number of clusters, we performed an LDA analysis using numbers varying from 5 to 30 for each analysis.

We utilized three evaluation metrics to compare the different numbers of topics, including topic coherence [29], topic similarity [30], and topic diversity [31]. Topic coherence measures the semantic similarity between high-scoring words in the topic, with higher values indicating better coherence. Topic similarity, calculated via the Jaccard index, assesses how similar two clusters are based on the words within the topics, with lower values indicating better similarity. Topic diversity is defined as the percentage of unique words across the top words of all topics. A diversity value closer to 0 implies redundant topics, while a value closer to 1 signifies more varied topics. A good number of topics are expected to have high topic coherence, high topic diversity, and low topic similarity. To determine the optimal K, we averaged the metric scores across the 10 iterations for a given K. The composite score was computed by taking a weighted average across the three metrics, assigning equal weights to topic coherence and diversity, and a negative weight to topic similarity to penalize redundancy. The K with the highest composite score was chosen as the optimal solution.

We further examined the top three iterations with the highest performance under the selected K. Two domain experts independently reviewed the semantics of each topic, merged subtopics with overlapping meanings based on proportional contribution, and assigned descriptive labels that best represented the content. A similar procedure was applied at the word level. When disagreements arose among the experts, a consensus discussion with clinicians was held to reconcile differences. The overall workflow is shown in **Figure 1**. The code is publicly available at: https://github.com/uf-hobi-informatics-lab/HIV-Stigma-Topic-Modeling.

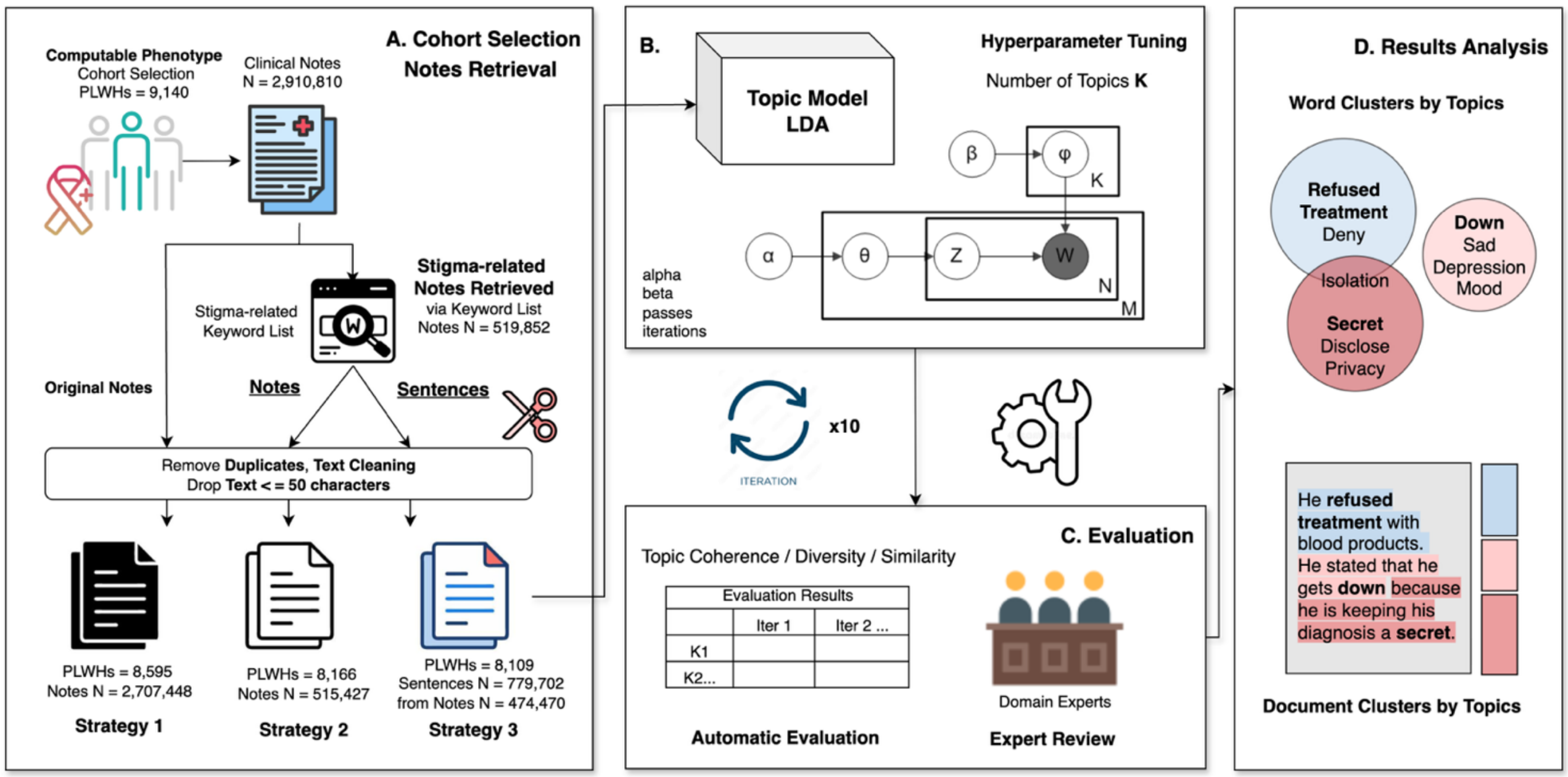

**Figure 1**. The overview of the topic modeling workflow.

## 3. RESULTS

### 3.1. Summary of PLWHs cohort and Clinical Notes

A total of 9,140 PLWHs were identified using the computable phenotype algorithm. **Table 1** summarizes demographics, clinical characteristics, and HIV-related measures. Observed comorbidities included substance use disorder, alcohol use disorder, mental health conditions, cardiovascular disease, sexually transmitted infections, and other chronic non-communicable diseases. At the most recent visit, 32.6% of patients had viral load suppression, and 77.1% had documented antiretroviral therapy (ART) prescriptions. **Figure 2** shows patient distributions by age, sex, race/ethnicity, and clinical characteristics, as well as the most common clinical note types, including progress notes (37.3%), emergency department notes (15.7%), and telephone encounters (11.1%).

After applying the snowball strategy, extending synonyms via word embeddings, and conducting manual review by clinicians, the final keyword list contains a total of 91 words. Using the list, we identified 2,707,448 notes, 515,417 notes, and 779,702 sentences from 474,470 unique notes for Strategies 1, 2, and 3, respectively. A detailed flowchart of selected patient and clinical notes information is shown in **Figure 1A**.

**Table 1**. Characteristics of people living with HIV identified from UF Health, n = 9,140.

| | Characteristics | N | % |
|---|---|---|---|
| **Age at first HIV encounter** | | | |
| | <18 | 119 | 1.3 |
| | 18-35 | 2681 | 29.3 |
| | 36-65 | 5961 | 65.2 |
| | 65+ | 379 | 4.1 |
| **Sex** | | | |
| | Female | 3502 | 38.3 |
| | Male | 5638 | 61.7 |
| **Race and Ethnicity** | | | |
| | Hispanic | 516 | 5.6 |
| | Non-Hispanic Black | 5494 | 60.1 |
| | Non-Hispanic White | 2826 | 30.9 |
| | Other | 185 | 2.0 |
| | Unknown | 119 | 1.3 |
| **Primary Insurance** | | | |
| | Medicaid | 1652 | 18.1 |
| | Medicare | 1544 | 16.9 |
| | Private | 1090 | 11.9 |
| | Other governmental [a] | 325 | 3.5 |
| | Self-pay | 404 | 4.4 |
| | No payer specified/Unknown | 4125 | 45.2 |
| **Clinical Characteristics** | | | |
| | Alcohol use disorder | 1536 | 16.9 |
| | Substance use disorder [b] | 2529 | 27.7 |
| | Cardiovascular disease | 1001 | 11.0 |
| | Congestive heart failure | 1250 | 13.7 |
| | Myocardial infarction | 676 | 7.4 |
| | Diabetes | 1904 | 20.8 |
| | Bacterial STIs [c] | 1080 | 11.8 |
| | HPV | 967 | 10.6 |
| | Hepatitis B or C | 1631 | 17.8 |
| | Anxiety | 2281 | 25.0 |
| | Bipolar | 1045 | 11.4 |
| | Depression | 2717 | 29.7 |
| **Having suppressed viral load (<200 ml/copies) at most recent visit** | | | |
| | Yes | 2984 | 32.6 |
| | No | 1745 | 19.1 |
| | No records of HIV viral load | 4411 | 48.3 |
| **Receiving ART prescriptions** | | | |
| | Yes | 7044 | 77.1 |
| | No | 2096 | 22.9 |

a: Other governmental agencies include VA, Tricare, Ryan White, the Department of Corrections, etc. b: Substance use disorder includes opioids, cannabis, cocaine, sedatives, hallucinogens, and other unspecified substances use disorder. c: Bacterial sexually transmitted infections (STIs) include syphilis, gonorrhea, and chlamydia

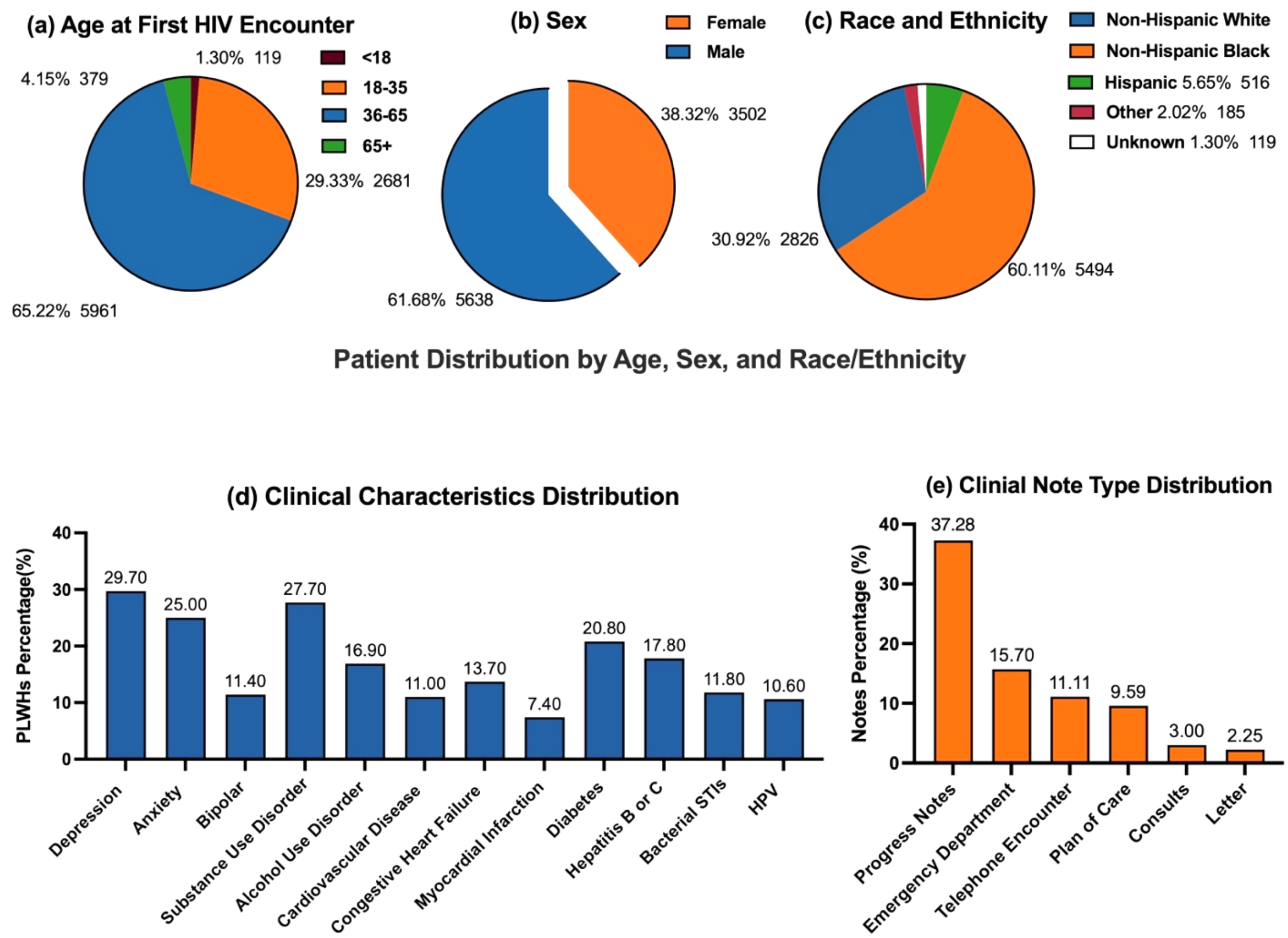

**Figure 2.** Patient distribution by age, sex, race/ ethnicity, and selected clinical characteristics. The figure also displays the distribution of the top six clinical note types in the dataset.

### 3.2. Optimized Number of Topics

We conducted topic modeling analysis using the three strategies described in the methods. For Strategy 1 (using all notes) and Strategy 2 (notes with at least one keyword), the optimized number of topics was identified as 25. For Strategy 3 (using sentences that contain at least one keyword), the optimized number of topics was identified as 15. Following the typical topic modeling analysis, we identified the top 10 words from the clinical notes according to the word-level probability to balance clarity and thoroughness for each topic. The summarized topics with the top 10 words for the three strategies are provided in Tables 2, 3, and 4.

**Table 2** presents 10 topics discovered using topic modeling from all notes from PLWHs (Strategy 1), ranging from lab results and medications to social context, comorbidities, physical exams, mental health, reproductive health, and administrative details. This creates a comprehensive overview of routine HIV management and the structural landscape of HIV care documentation.

**Table 2**. Topics identified in all clinical notes from PLWHs.

| | **Topics** | **Keywords** |
|---|---|---|
| 1 | Lab Results | lab, range, result, status, cholesterol, glucose, negative, component, blood, specimen |
| 2 | Medication | tablet, daily, visit, vaccine, medication, capsule, medicine, instruction, treatment, intravenous |
| 3 | History Medical Records | date, negative, HIV, history, visit, medication, result, vaccine, complaint, lesion |
| 4 | Social Records | social, impression, status, activity, oral, tobacco, sexual, medical, partner, CMSHCC |
| 5 | Patient Care | discharge, home, care, therapy, treatment, transfer, assessment, pain, monitor, bedside |
| 6 | Physical Examination | negative, pain, abdominal, skin, tenderness, breath, neck, chest, ear, radiology |
| 7 | Comorbidities and Side Effect | heart, cardiac, dialysis, anesthesia, catheter, complication, blood, hemodialysis, chest, fracture |
| 8 | Psychiatric Health, Substance Abuse, Mental Health | disorder, denies, depression, mood, abuse, suicidal, mental, anxiety, substance, mental |
| 9 | Reproductive Health | patient, negative, HIV, vaginal, pregnancy, nutrition, food, diet, age, pap |
| 10 | Appointment and Contacts Information | name, date, locationcity, locationzip, locationstreet, contactphone, account, appointment, physician, visit |

* The Centers for Medicare and Medicaid Services Hepatocellular Carcinoma (CMSHCC)

**Table 3** presents the topics discovered from notes that contain at least one stigma-related keyword (Strategy 2). We discovered important medical topics, such as medication adherence, testing, and musculoskeletal and radiology findings, and 5 novel stigma-specific topics, including mental health distress, disclosure concerns related to privacy and family fears, social activities encompassing community engagement and sexual behavior, healthcare bias reflecting differential treatment by providers, and lifestyle factors shaped by stigma.

**Table 3**. Topics identified from clinical notes that contain at least one description of stigma.

| | Topics | Keywords |
|---|---|---|
| 1 | Mental Health Distress | disorder, history, denies, depression, mood, abuse, suicidal, mental, anxiety, substance |
| 2 | Disclosure Concern | denies, pain, lesion, status, history, family, normal, negative, check, privacy |
| 3 | Social Activities | oral, activity, history, social, sexual, abused, partner, organization, attends, club |
| 4 | Healthcare Biased | differential, viral, infection, manual, below, appointment, discharge, physician, complaint, impression |
| 5 | Lifestyle Factors | health, food, care, pressure, high, help, drink, heart, diet, exercise |
| 6 | Preventive Care | vaccine, medication, health, current, risk, screening, maintenance, status, pain, normal |
| 7 | HIV Testing | negative, blood, detected, specimen, PCR, virus, susceptible, smear, glucose, neutrophil |
| 8 | Musculoskeletal Issue | pain, knee, fracture, joint, back, leg, spine, shoulder, ankle, lumbar |
| 9 | Radiology Reports | finding, right, left, date, image, chest, CT, electronically, lung, above |
| 10 | Medication | tablet, take, mouth, daily, capsule, needed, medication, bedtime, nightly, solution |
| 11 | Cardiovascular/Renal | dialysis, heart, renal, HD, ESRD, chronic, systolic, failure, ventricular, cardiovascular |

* Polymerase chain reaction (PCR); Hemodialysis (HD); End-stage renal disease (ESRD)

**Table 4** provides the 11 topics discovered using only stigma-related sentences (Strategy 3) along with the associated top 10 words, identified from the 15 topics discovered through topic modeling. These 11 summarized topics cover a wide range of challenges faced by PLWHs. The topic "Mental Health Concerns and Stigma" highlights critical issues such as depression, anxiety, and suicidal ideation, illuminating the psychological burdens and societal stigma that individuals navigate. Topics "Social Support and Engagement", "Limited Healthcare Access and Severe

Illness," and "Food Insecurity and Resource Scarcity" emphasize the obstacles to access to adequate healthcare and essential social resources. Topics including "Missed Appointments and HIV Care Monitoring" and "Treatment Refusal and Isolation" reflect the practical difficulties associated with sustaining consistent healthcare treatment. Additionally, the topic "Intimate Partner Violence and Relationship Issues" reveals the complexities of interpersonal relationships and the emotional and physical risks that may arise. The topic "Substance Abuse" captures alcohol, drug, and other substance use among PLWHs, which could potentially impair judgment, heighten the risk of HIV transmission, and exacerbate the overall health consequences of HIV. Topics "Fear of Falling and Physical Health Concerns", "Medical History and Screening," and "Physical Symptoms and HIV Pain Management" capture information about symptom management in improving the quality of life for PLWHs.

**Table 4**. Topics identified in sentences containing descriptions of stigma.

| | **Topics** | **Keywords** |
|---|---|---|
| 1 | Mental Health Concerns and Stigma | denies, feeling, depression, mood, poor, anxiety, suicidal, sad, sleep, loss |
| 2 | Medical History and Screening | history, past, medical, diagnosis, exam, result, disorder, procedure, test, screening |
| 3 | Social Support and Engagement | attends, organization, club, partner, abused, stress, family, friend, social, frequency, |
| 4 | Limited Healthcare Access and Severe Illness | lack, evaluation, contrast, HIV, limited, intravenous, offer, illness, onset, severe |
| 5 | Missed Appointments and HIV Care Monitoring | appointment, medication, missed, daily, cancel, scheduled, visit, negative, outpatient, clinic |
| 6 | Treatment Refusal and Isolation | refused, isolation, declined, treatment, emotional, problem, risk, intervention, precaution, surgery |
| 7 | Intimate Partner Violence and Relationship Concerns | abused, partner, relationship, sexual, fear, physically, violence, club, intimate, concern |
| 8 | Fear of Falling and Physical Health Concerns | fall, fear, patient, exercise, need, activity, assessment, risk, barrier, effective |
| 9 | Substance Abuse | pain, use, drug, alcohol, substance, tobacco, activity, management, control, abdominal |
| 10 | Food Insecurity and Resource Scarcity | transportation, food, medical, lack, social, need, stress, insecurity, physical, financial |
| 11 | Physical Symptoms and HIV Pain Management | bad, therapy, temp, discharge, hurt, infection, muscle, disease, headache, symptom |

We discovered more HIV stigma dimensions, social, and related behavioral circumstances topics by filtering the input text. The topic modeling analysis using all notes discovered 2 related topic out of 10 topics (20%), "Social Records" and "Psychiatric Health, Substance Abuse, Mental Health"; the analysis using notes that contain at least one keyword discovered 5 related topics out of 11 topics (around 45%), including "Mental Health Distress", "Disclosure Concern", "Social Activities", "Healthcare Bias", and "Lifestyle Factors"; the topic analysis using only sentences contain at least one keyword 9 related topics out of 11 topics (around 82%), including "Mental Health Concern and Stigma", "Social Support and Engagement", "Limited Healthcare Access and Severe Illness", "Missed Appointments and HIV Care Monitoring", "Treatment Refusal and Isolation", "Intimate Partner Violence and Relationship Concerns", "Fear of Falling and Physical Health Concerns", "Substance Abuse", and "Food Insecurity and Resource Scarcity".

### 3.3. Word Frequency Analysis

Word frequency analysis is a widely used algorithm to generate a word cloud to highlight the popular words associated with individual topics. Word frequency analysis offers detailed information about the topics to add more details to topic modeling. Our word frequency analysis focused on the results from Strategy 3, as it captures more topics of stigma dimensions, social, and related behavioral circumstances.

**Figure 3** shows a word cloud illustrating the top 10 most common words across 11 topics discovered using Strategy 3. The word cloud represents topics generated from the entire document, with different colors representing distinct topics and word size indicating the proportion of text associated with topics, proportional to the density *p (word | topic)*. Key topics include “Food”, “Transportation”, “Lack”, “Refused”, “Appointment”, and “Medication”, which highlight barriers to care, such as missed appointments, poor adherence to treatment, and limited access to healthcare. Mental health challenges are reflected in terms like “Depression”, “Anxiety”, “Suicidal”, and “Isolation”, emphasizing the stigma and the impact of fears regarding confidentiality and societal rejection on patients' experiences. Additionally, words such as “Pain”, “Muscle”, “Drug”, and “Abuse” underscore the physical and substance-related challenges faced by PLWHs.

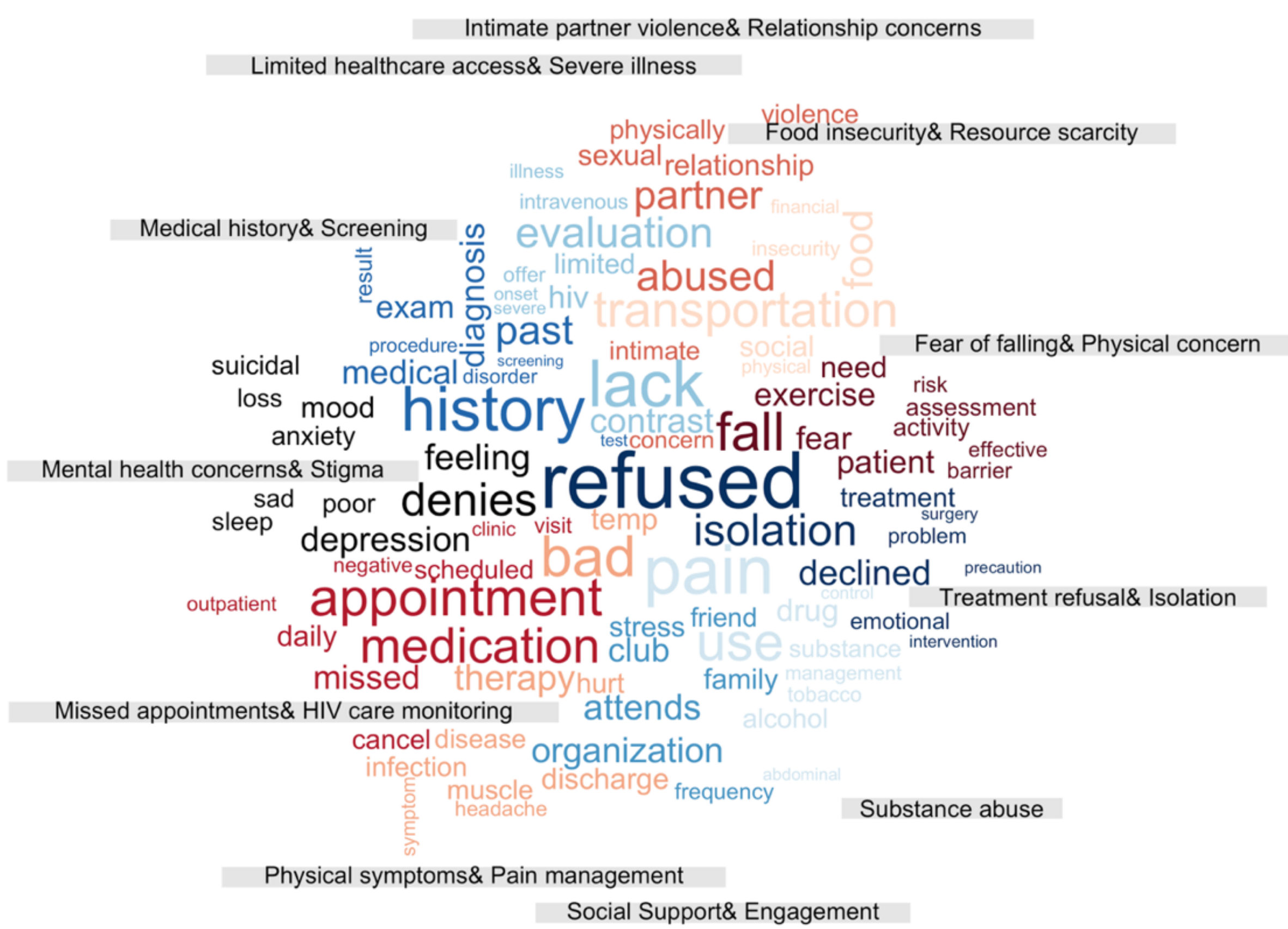

**Figure 3**. Word cloud. The top ten keywords from each of the eleven LDA topics were extracted from stigma-related sentences. The size of each word indicates its frequency, while each color represents a different topic. Grey-shaded labels indicate the names of the topics.

### 3.4. Topic Variance among Sex and Age-stratified Groups

Following Griffiths and Steyvers et al. [32], we used a document-topic matrix to capture the probabilities of various topics for all documents. To investigate potential sex differences, we used the 'Sex' feature derived from structured EHR, calculated sex-specific topic probabilities by averaging the topic distribution across all documents from male and female patients, respectively. As shown in **Figure 4a**, the heatmap cell color intensity indicates a higher proportion of sentences in that topic assigned to each group. Overall, the average probabilities of topics are not remarkably different between the male and female groups. The absolute differences in prevalence between male and female patients were generally <0.01 across all topics. Both groups shared the same rank ordering of topic prevalence. These findings suggest that similar information was documented for the two sex groups.

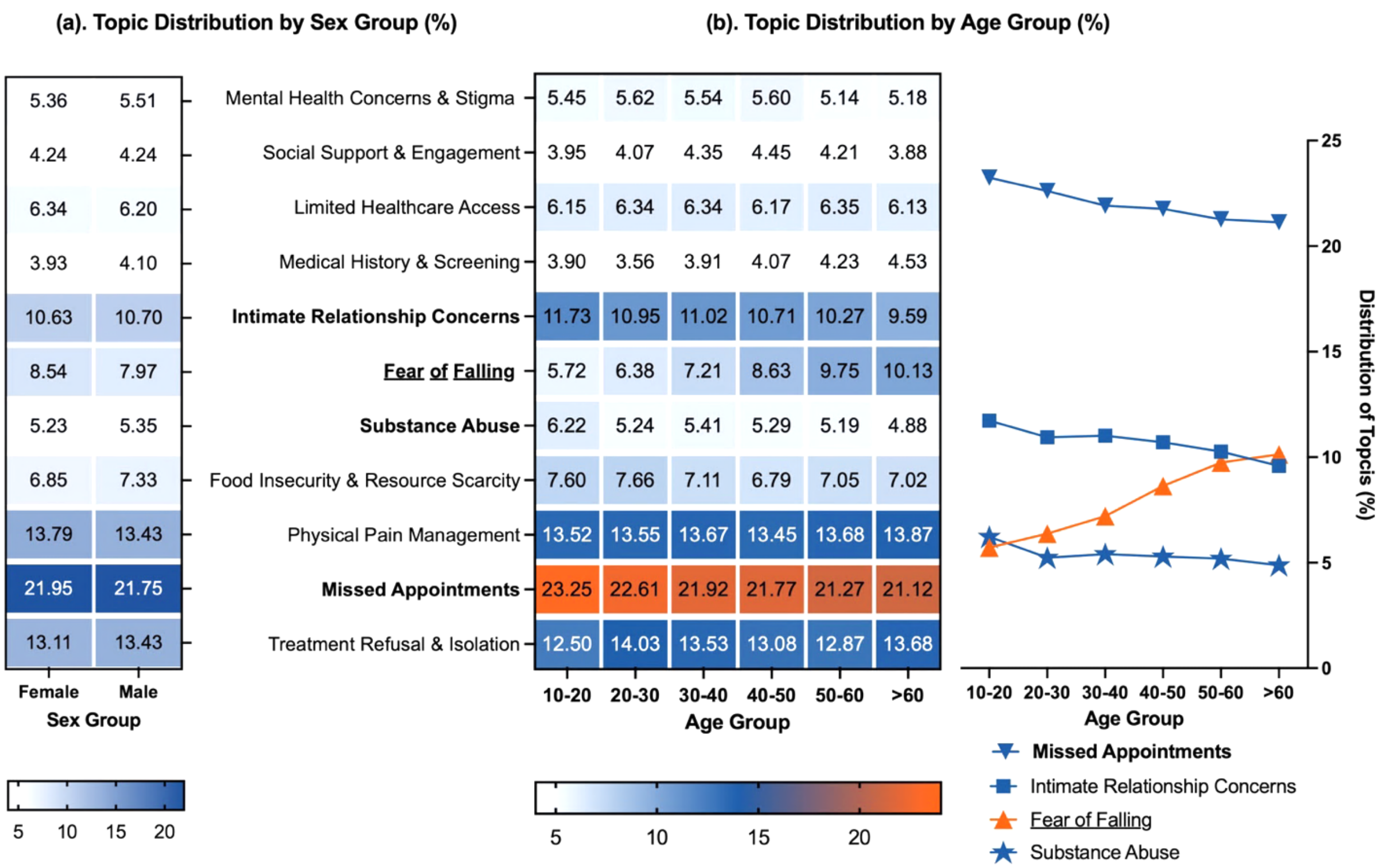

**Figure 4**. Topic distribution across sex and age groups.

Similar to the sex-specific analysis, we stratified the cohort into six age-specific groups ranging from 10 to 90 years old. For each age group, we calculate the average probabilities of topics and compare the differences across all age groups. **Figure 4b** shows the comparison results. The variations of probability among age groups are remarkably larger than the gender-based groups (> 0.01). Among the age groups, three topics show consistent decline in probability as age increases, including “Substance Abuse”, “Intimate Partner Violence and Relationship Concerns”, and “Missed Appointments and HIV Care Monitoring”; one topic, “Fear of Falling and Physical Health Concerns”, shows a remarkable increase in probability in older age groups. There are no remarkable differences for other topics across age groups, such as “Social Support and Engagement”, “Medical History and Screening”, “Food Insecurity and Resource Scarcity”, and “Limited Healthcare Access and Severe Illness”.

## 4. DISCUSSION AND CONCLUSION

In this study, we applied topic modeling to analyze around 2.9 million clinical notes from PLWHs receiving care at a large academic medical center in Florida to discover topics of stigma dimensions, social, and related behavioral circumstances. Using an unsupervised topic modeling based on LDA, we experimented with three different strategies and discovered related topics buried in the large numbers of clinical notes. The topic analysis using Strategy 3, followed by manual review and summarization, generated 11 topics capturing medical, social, behavioral, and stigma information. The following sex and age-specific analysis further examined the topic differences among sex and age groups. This study discovered important topics captured in narrative clinical notes for PLWHs to facilitate future studies to perform a deep analysis of clinical notes for PLWHs.

The stigma-related dimensions and related social and behavioral contexts topics identified in our study aligned with findings from previous studies. Topics such as "Food Insecurity and Resource Scarcity" and "Healthcare Access Barriers" are important socioeconomic factors associated with health outcomes of PLWHs, with studies showing that enhancing food security and strengthening social and healthcare support can benefit both mental health and HIV outcomes in resource-limited settings [9], [33]. The topic "Mental Health Concerns and Stigma" aligns with existing findings that high rates of HIV stigma, depression, and past trauma in PLWHs, and about half of these individuals are under treatment [6], [34]. The topic "Substance Abuse" captures a higher level of drug dependence and other substance use among PLWHs compared to the HIV-uninfected population [35]. Both topics align closely with cohort clinical characteristics. The topic "Physical symptoms and HIV pain management" and "Fear of Falling and Physical Health Concerns" often relates to musculoskeletal pain and falling experienced by PLWHs, which remains common even after patients achieve near-normal immune function [36], [37]. The sex- and age-specific analysis using the average probabilities revealed no notable differences in topics among sex groups. However, differences were observed in age groups for certain topics, aligning with prior studies indicating that the older group tends to have lower substance use, fewer social contacts, less visit missingness, and more physical pain [38], [39].

Understanding the social, behavioral, and clinical context of HIV stigma is the very first step for future studies to examine the details and develop early interventions to improve treatment outcomes and quality of life for PLWHs. There are many strengths in our study. We targeted HIV-related sensitive psychosocial domains that have not previously been characterized from a large number of real-world EHR notes by topic modeling. Guided by domain experts, we iteratively developed the stigma keyword list and experimented with three different filtering strategies systematically. The third strategy identified the sentences using a stigma-related keyword list, which was able to identify more topics of interest. We experimented with different numbers of topics and evaluated results using both qualitative (manual review and summarization by experts) and quantitative (topic coherence, topic similarity, and topic diversity) evaluation metrics. To better interpret the topics, we performed word frequency analysis using a word cloud and examined the topics across different sex and age groups using structured EHR information to provide insights into the differences among subgroups of PLWHs. Although our study focused on HIV-related stigma, the proposed method can be applied to other disease domains (e.g., hepatitis, cancer) and other healthcare institutions with minimal effort to customize the disease-specific keywords.

This study has limitations. We examined the EHR from a single healthcare institution, as privacy protections and the lack of public HIV cohorts with clinical notes limit multi-institutional access.  We manually summarized topic labels based on each cluster, which may not reflect the content outside of these clusters. We also acknowledge that topic modeling is an unsupervised method, and by design, it may underrepresent rare themes. Individuals with more visits may still contribute more notes to the dataset, which could potentially skew the topic distribution. In our future research, we plan to develop natural language processing tools to capture more details of these topics and study the role of stigma dimensions, social, and related behavioral circumstances in HIV outcomes across multiple healthcare systems.

## SUMMARY TABLE

**What was already known on the topic**:

- People living with HIV are known to experience stigma, triggering mental health challenges and undermining treatment adherence and health outcomes.
- Social and related behavioral circumstances discourage HIV-infected people from seeking help, negatively impacting their mental health

**What this study added to our knowledge**:

- The first large-scale topic modeling study on the documentation of stigma, social, and behavioral factors in clinical narratives for people living with HIV
- Discovers 11 topics from large-scale clinical narratives to improve the understanding of documentation of stigma, social, and health behaviors for people living with HIV
- Age and sex group analysis discovers differences in the documentation of stigma, social, and health behaviors related information.

## SEX-BASED ANALYSES

This study performed sex-based subgroup analysis to examine the differences in documenting stigma-related topics using the sex information captured in structured electronic healthcare records.

## ACKNOWLEDGMENTS

This study was partially supported by grants from the National Institute of Allergy and Infectious Diseases, NIAID R01AI172875 and National Institute of Mental Health, NIMH R34MH135768, Patient-Centered Outcomes Research Institute® (PCORI®) Award (ME-2018C3-14754, ME-2023C3-35934), the PARADIGM program awarded by the Advanced Research Projects Agency for Health (ARPA-H), National Institute on Aging, NIA R56AG069880, National Heart, Lung, and Blood Institute, R01HL169277, National Institute on Drug Abuse, NIDA R01DA050676, R01DA057886, the National Cancer Institute, NCI R37CA272473, and the UF Clinical and Translational Science Institute. The content is solely the responsibility of the authors and does not necessarily represent the official views of the funding institutions. We gratefully acknowledge the support of NVIDIA Corporation and the NVIDIA AI Technology Center (NVAITC) UF program.

## COMPETING INTERESTS STATEMENT

Ziyi Chen, Yiyang Liu, Mattia Prosperi, Krishna Vaddiparti, Robert L Cook, Jiang Bian, Yi Guo, and Yonghui Wu have no conflicts of interest that are directly relevant to the content of this study.

## CONTRIBUTORSHIP STATEMENT

ZC, MP, and YW were responsible for the overall design, development, and evaluation of this study. ZC was involved in the analysis of the results. ZC and YW did the initial drafts. ZC, YL, MP, KV, RC, JB, YG, and YW did the revisions of the manuscript. All authors reviewed the manuscript critically for scientific content, and all authors gave final approval of the manuscript for publication.